%% file: main.tex
\documentclass{article} 

\usepackage{iclr2027_conference, times}

\iclrfinalcopy

\usepackage[utf8]{inputenc} 
\usepackage[T1]{fontenc}    
\usepackage{url}            
\usepackage{booktabs}       
\usepackage{amsfonts}       
\usepackage{nicefrac}       
\usepackage{microtype}      
\usepackage{xcolor}         

\usepackage{tikz}
\usetikzlibrary{calc}

\usepackage{colortbl}

\usepackage{amsmath}
\usepackage{amssymb}
\usepackage{mathtools}
\usepackage{amsthm}

\usepackage{wrapfig}
\usepackage{listings}
\usepackage{tabulary}
\usepackage{etoolbox}
\usepackage{multirow}
\usepackage{enumitem}
\usepackage{xspace}

\usepackage{tikz}
\usepackage{pgfplots}
\usepackage{natbib}

\usepackage{algorithm,algorithmicx,algpseudocode}
\usepackage{subcaption}

\usepackage[pagebackref=true,breaklinks=true,colorlinks,urlcolor={red!90!black},linkcolor={red!90!black},citecolor={blue!90!black},bookmarks=false]{hyperref}
\usepackage[toc,page,header]{appendix}
\usepackage{minitoc}

\title{Linear Reusable Neural Bases Architecture for Network Compression}

\author{Binshuai Wang, Peng Wei \\
Department of Computer Science \\
The George Washington University \\
Washington D.C., USA \\
\texttt{\{derekwang, pwei\}@gwu.edu} \\
\AND
Mahyar Ghazanfari \\
Department of Mechanical and Aerospace Engineering \\
The George Washington University \\
Washington D.C., USA \\
\texttt{\{Mahyar.ghazanfari\}@gwu.edu} \\
}

\input{math_commands}

\begin{document}

\maketitle

\begin{abstract}
Memory constraints remain a critical bottleneck in the deployment of large-scale AI models. 
Parameter sharing across network depth reduces model storage, but repeatedly applying an identical transformation limits flexibility across layers. 
Inspired by time--memory trade-offs in classical algorithms, we introduce the Linear Reusable Neural Bases (LRNB) architecture, an RNN-based framework that improves parameter efficiency through parameter reuse at the \textit{neuron level}. 
Each feedforward residual module is represented as a linear combination of shared neural bases, with depth-specific learnable coefficients and optional shifts providing flexibility across layers.
This formulation reduces parameter redundancy across depth and enables the construction of wider and deeper networks within a fixed parameter budget. 
We further provide a geometric interpretation of the neural bases from a vector-field perspective and extend the framework to linear projection modules. 
Experiments demonstrate that the LRNB architecture achieves comparable or lower final training loss than independently parameterized baselines while using fewer parameters and maintaining stable training dynamics. 
These findings support neuron-level reuse as a practical approach to parameter-efficient network design.
\end{abstract}



\input{sec/intro}

\input{sec/bg}

\input{sec/method}

\input{sec/disc}

\input{sec/exp}

\input{sec/con}

\bibliographystyle{iclr2027_conference}
\bibliography{myBib}

\appendix 

\input{sec/add_nbs}

\input{sec/add_exps}

\end{document}

%% file: math_commands.tex
\usepackage{amsmath,amsfonts,bm}

\def\eqref#1{equation~\ref{#1}}

\def\1{\bm{1}}

\DeclareMathAlphabet{\mathsfit}{\encodingdefault}{\sfdefault}{m}{sl}
\SetMathAlphabet{\mathsfit}{bold}{\encodingdefault}{\sfdefault}{bx}{n}



%% file: sec/intro.tex
\section{Introduction}
\label{sec:intro}
large-scale AI models outperform smaller ones across a broad range of domains~\citep{Brown2020gpt3,hoffmann2022training} and have attracted significant attention in the machine learning community.
As indicated by empirical scaling laws~\citep{Kaplan20scalinglaws, hoffmann2022training}, models with a larger parameter count typically have stronger approximation capabilities.
Consequently, modern AI models are often designed with billions or even trillions of parameters, requiring hundreds or thousands of GPUs/TPUs for training and deployment~\citep{Brown2020gpt3,Fedus2021switch}.

Although large-scale AI models have demonstrated remarkable performance, deployment remain challenging, especially for researchers and organizations with limited computational resources.
One of the major bottlenecks is the memory requirement.
Deploying a large-scale model often requires operating hundreds or thousands of GPUs simultaneously, and the high cost of specialized hardware further increases the financial burden~\citep{Narayanan2021megatron}.
Overall, these resource and hardware demands hinder many researchers and organizations from deploying large-scale AI models.

A promising approach to addressing this challenge is to integrate a neuron-reuse mechanism into the model, to reduce parameter redundancy and memory requirements.
This idea is closely related to recurrent neural network (RNN) models, which repeatedly apply a shared transformation across different time steps or positions~\citep{Elman90Finding,JORDAN97Serial}. 
Moreover, consistent with the philosophy that \textit{compression leads to intelligence}, models representing information with fewer parameters often have better generalization.
Nevertheless, classical RNN models often suffer from optimization difficulties, as repeated transformations can result in unstable gradient propagation, such as vanishing or exploding gradients. 
As a result, most large-scale AI models are designed with independent modules and integrate only limited reuse mechanisms.

In light of these observations, a natural question arises:

\textit{Can we design an architecture that integrates the reuse mechanism at the neuron level to improve parameter efficiency and lower memory costs, while maintaining training stability?}

In this paper, we introduce a novel model architecture, the Linear Reusable Neural Bases (LRNB) architecture, to improve parameter efficiency through parameter reuse at the \textit{neuron level}.
Inspired by sinusoidal bases in Fourier analysis and linear bases in linear algebra, we introduce the concept of \textit{neural bases}, with an explanation from a vector-field perspective.
We represent each neural network module as a linear combination of shared neural bases with allowable shifts, enabling the same neurons to be reused across depths. 
This reuse mechanism reduces the parameter redundancy along the depth dimension and improves the model compression rate, allowing for the construction of substantially wider and deeper networks under the same parameter budget. 
Experimental results show that our architecture achieves comparable or lower final loss than classical architectures, while maintaining stable training dynamics and reducing the parameter count.

\paragraph{Organization.}
The remainder of this paper is organized as follows.
Section~\ref{sec:related} reviews related work.
Section~\ref{sec:bg} provides background on deep learning theory and residual networks.
Section~\ref{sec:method} introduces the theory of neural bases, the model architecture, and the implementation details.
Section~\ref{sec:disc} discusses the advantages and limitations of the model.
Section~\ref{sec:exp} presents computational experiments to evaluate the model.

\paragraph{Notation.}
We use $\sigma(\cdot)$ to denote a nonlinear activation function, $\phi(\cdot)$ a neural basis function, and $R(\cdot)$ a residual block.
The notation $\langle \cdot,\cdot\rangle$ denotes the inner product.
We write $d$ for the embedding dimension, $m$ for the network width (i.e., the hidden dimension or number of neurons), $L$ for the network depth, and $T$ for the token sequence length.
For a residual feedforward block, $(S,b)$ denotes the sensor parameters, and $P$ denotes the response parameters.
For the $j$-th neural basis function (or neuron), the corresponding sensor and response parameters are denoted by $(s_j,b_j)$ and $p_j$, respectively.

\vspace{-0.2in}
\begin{figure}[ht]
    \centering
    \begin{minipage}[c]{0.66\textwidth}
        \centering
        \includegraphics[
        width=\linewidth,
        trim={20pt 20pt 40pt 15pt},
        clip]{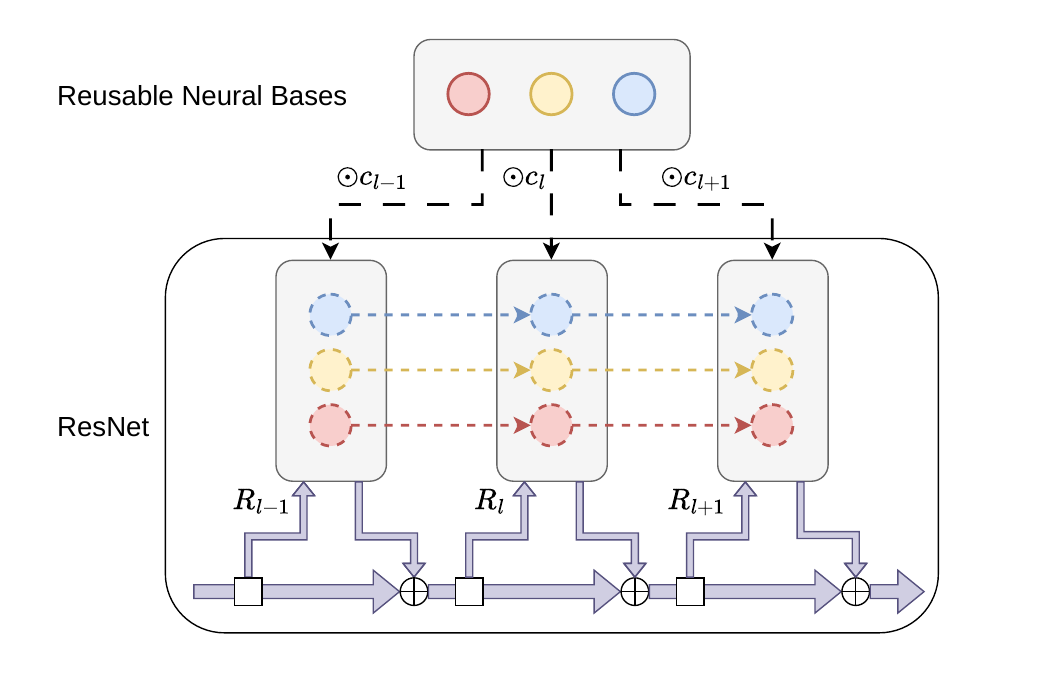}
    \end{minipage}
    \begin{minipage}[c]{0.3\textwidth}
        \vspace{0.15in}
        \caption{
        Illustration of the linear reusable neural bases architecture.
        Residual modules $R_{l-1}$, $R_l$, and $R_{l+1}$ are constructed from a shared set of reusable neural bases (matching colors), with depth-specific coefficient vectors $c_{l-1}$, $c_l$, and $c_{l+1}$ determining their respective contributions.
        Colored dashed arrows indicate neuron-reuse mechanism across depths.
        Purple arrows show feature flow through the ResNet.
        The symbols $\odot$ and $\oplus$ denote element-wise
        multiplication and residual addition, respectively.
        }
        \label{fig:lrnba_arch}
    \end{minipage}
\end{figure}
\vspace{-0.4in}

\section{Related Work}
\label{sec:related}

Our work relates to three complementary lines of research: recurrent architectures, structured and shared parameterizations, and parameter sharing in Transformers. We review these directions and then clarify how our approach differs in the granularity of reuse.

\paragraph{Recurrent Architectures.}
Recurrent neural networks (RNNs) provide a classical mechanism for parameter reuse by repeatedly applying shared transformations to evolving hidden states.
Representative architectures include Elman networks, Jordan networks, long short-term memory networks (LSTMs), and gated
recurrent units (GRUs)~\citep{Elman90Finding, JORDAN97Serial,
Hochreiter97LSTM, Cho2014Learning}.
Such sharing reduces the number of independently learned
parameters, but repeated composition can lead to vanishing
or exploding gradients during backpropagation through
time~\citep{Bengio1994long, Hochreiter97LSTM,
Pascanu2013difficulty}.
LSTMs and GRUs mitigate these difficulties through gating,
while other recurrent architectures explore orthogonal
transformations~\citep{helfrich2018orthogonal} and
convolutional structures with connections to recurrent
computation~\citep{bai2018trellis}.
These approaches motivate the joint consideration of
parameter reuse and optimization stability.

\paragraph{Structured and Shared Parameterizations.}
Another line of research improves parameter efficiency
by imposing structure on network weights or constructing
them from shared components.
Low-rank approaches compress recurrent
transformations~\citep{Lu16compact}, while tensorized neural
networks represent dense transformations using tensor-train
decompositions~\citep{novikov2015Tensorizing}.
Related methods employ structured matrices to reduce
parameter storage~\citep{Sindhwani15structured}.
Beyond factorization, HyperNetworks use auxiliary networks
to generate the weights of a target
network~\citep{ha2016hypernetworks}.
Residual adapters combine shared representations with
task-specific parameters~\citep{Rebuffi17adapters},
whereas shared-template methods construct layer weights
as learned linear combinations of common parameter
templates~\citep{savarese18parametersharing}.
Basis-based sharing~\citep{wang2025basis} is also closely
related to our approach, although it operates at the
layer level rather than directly reusing neuron-level
nonlinear features.

\paragraph{Parameter Sharing in Transformers.}
Parameter sharing has also been extensively explored
in Transformer architectures~\citep{Vaswani17attention}.
ALBERT shares parameters across encoder layers to reduce
the number of independently learned
weights~\citep{lan2020albert}.
The Universal Transformer repeatedly applies shared
self-attention and transition blocks across
depth~\citep{dehghani2019universal}, and the Relaxed
Recursive Transformer explores a relaxed form of recursive
parameter sharing~\citep{bae2025relaxed}.
The Sparse Universal Transformer combines depth-wise
sharing with sparse mixture-of-experts
layers~\citep{tan2023sparseuniversaltransformer}.
Sharing can also be restricted to particular components:
\citet{Pires23Onewide} investigate redundancy among
feed-forward network (FFN) blocks and share a single
wide FFN across multiple layers.
Closely related looped Transformers repeatedly apply
a fixed block, or a small set of blocks, to an evolving
hidden representation~\citep{giannou2023looped,
yang2024looped}.
These architectures connect depth-wise parameter reuse
with iterative computation and the learning of
algorithmic tasks.

\paragraph{Relation to Our Work.}
The LRNB architecture draws on recurrent computation and shared-basis parameterizations, but emphasizes reuse at the \textit{neuron level}.
Rather than sharing an entire transformation across depth or combining modules or sub-modules at the weight-matrix level, our approach constructs modules as learned linear combinations of shared nonlinear neural bases. The resulting architecture aims to improve parameter efficiency with flexibility across depth.

%% file: sec/bg.tex
\section{Background}
\label{sec:bg}

Before presenting our main results, we briefly review several key concepts and formulations.

\subsection{Deep learning}
Deep learning theory suggests that the approximation ability of neural networks can be improved by increasing the depth of sequentially composed architectures~\citep{lecun2015deep}.
By hierarchically composing nonlinear transformations, deep neural networks can represent certain functions more efficiently than the shallow networks~\citep{Raghu17Expressive}.
Consequently, deep networks often exhibit better approximation ability than the shallow networks for complex learning tasks.

Although the sequentially composed architecture of deep neural networks enables models to learn more complex features and representations, it also introduces higher computational and memory costs during training or inference, compared to shallow networks~\citep{Rajbhandari20zero}.
Since each module must be processed sequentially, a large number of parameters and intermediate activations must be retained in memory, even though only a small group of parameters are used in execution during training. 
This memory requirement grows with network depth, posing a significant burden for researchers with limited computational resources.

\subsection{Residual Networks}
Residual networks (ResNets) play an important role in deep learning by alleviating the vanishing-gradient problem and enabling the stable training of very deep networks~\citep{He2015res}.
The key feature of a ResNet is the use of shortcut connections, which allows the input information to be passed directly to deeper layers rather than propagating only through nonlinear layers.
\cite{Chen18NeuralODE} further provide a dynamical-systems perspective by interpreting residual updates as discretized flows of vector fields.

For example, a residual feedforward module in a ResNet can be written as
\[
x + R(x),
\]
where $x$ represents the shortcut connection and $R(x)$ represents the residual component.
A typical residual component takes the form
\begin{equation*}
\label{eq:resnet}
R(x)=P\sigma(Sx+b),
\end{equation*}
where $\sigma(\cdot)$ is an element-wise nonlinear activation function, and $(S,b)$ are the learnable parameters of the input affine layer, and $P$ are the learnable parameters of the output linear layer.\footnote{For simplicity, we omit the bias term in the output layer. This omission does not affect the nonlinear structure of the residual component, since the output bias contributes only a constant translation.}

From a more detailed perspective, the residual component can be viewed as a linear combination of a shared collection of nonlinear functions.
The parameters $(S,b)$ determine the directions and intercepts (offsets) of these nonlinear functions in the input space, while $P$ specifies the coefficients used to combine them in the output space.
We further extend this viewpoint in Section~\ref{sec:method} to develop a new model architecture that reuses parameters to reduce memory costs.



%% file: sec/method.tex
 \section{Methodology}
\label{sec:method}
In this section, we introduce our new framework, the Linear Reusable Neural Bases (LRNB) Architecture, designed to improve parameter efficiency and reduce memory cost.
The key idea is to reuse a shared group of neural bases while assigning different coefficients to construct modules at different positions, enabling the same neurons to be reused across depths.
In the remainder of this section, we first define neural bases and provide a geometric interpretation from the perspective of vector fields. 
We then describe the model architecture, training strategy, and key implementation details.

\subsection{The Neural Bases}
\label{subsec:nb}
In this subsection, we first introduce our core concept of neural bases and provide a geometric interpretation from the perspective of vector fields. 

\paragraph{Definition of Neural Bases.}
As introduced in Section~\ref{sec:bg}, a residual feedforward block has the form
\begin{equation*} 
     x + P \sigma(Sx + b).
\end{equation*}
Let $m$ denote the hidden dimension of the residual block, and write the rows of $S$, the entries of $b$, and the columns of $P$ as
\[
S^\top = [s_1,s_2,\ldots,s_m],
\qquad
b^\top = [b_1,b_2,\ldots,b_m],
\qquad
P = [p_1,p_2,\ldots,p_m],
\]
where $s_j^\top$ is the $j$-th row of $S$, $b_j$ is the $j$-th entry of $b$, and $p_j$ is the $j$-th column of $P$.

For each $j = 1, 2, \dots, m$, we define the function $\phi_j(x)$ as:
\[
\phi_j(x) := \sigma(\langle x, s_j \rangle + b_j) p_j.
\]
Then the residual component of the network can be rewritten as a summation of such functions:
\begin{equation*}
P \sigma(Sx+b)
=
\sum_{j=1}^m \sigma(\langle x,s_j\rangle + b_j)p_j
=
\sum_{j=1}^m \phi_j(x).
\end{equation*}
This formulation reveals that the nonlinear function of the residual block can be further decomposed into a group of parallel basic nonlinear functions $\{\phi_j\}_{j = 1}^m$.
Based on this decomposition, and since each of these functions serves as an \textit{atomic} nonlinear transformation from the input to the output space—in the sense of being the simplest and independent building unit—we refer to each function $\phi_j$ as a \textit{neuron} or a \textit{neural basis}\footnote{We use ``basis'' informally to denote an elementary and independent building unit, analogous to sinusoidal bases in Fourier analysis and basis vectors in linear algebra. However, orthogonality is not assumed.} in the network.
Moreover, we refer to the associated parameters $(s_j, b_j)$ for the input space as the \textit{sensor} parameter of neuron $\phi_j$ and $p_j$ as the \textit{response} parameter of neuron $\phi_j$.

\begin{figure}[ht]
    \centering
    \begin{subfigure}[t]{0.45\textwidth}
        \centering
        \includegraphics[width=\linewidth]
        {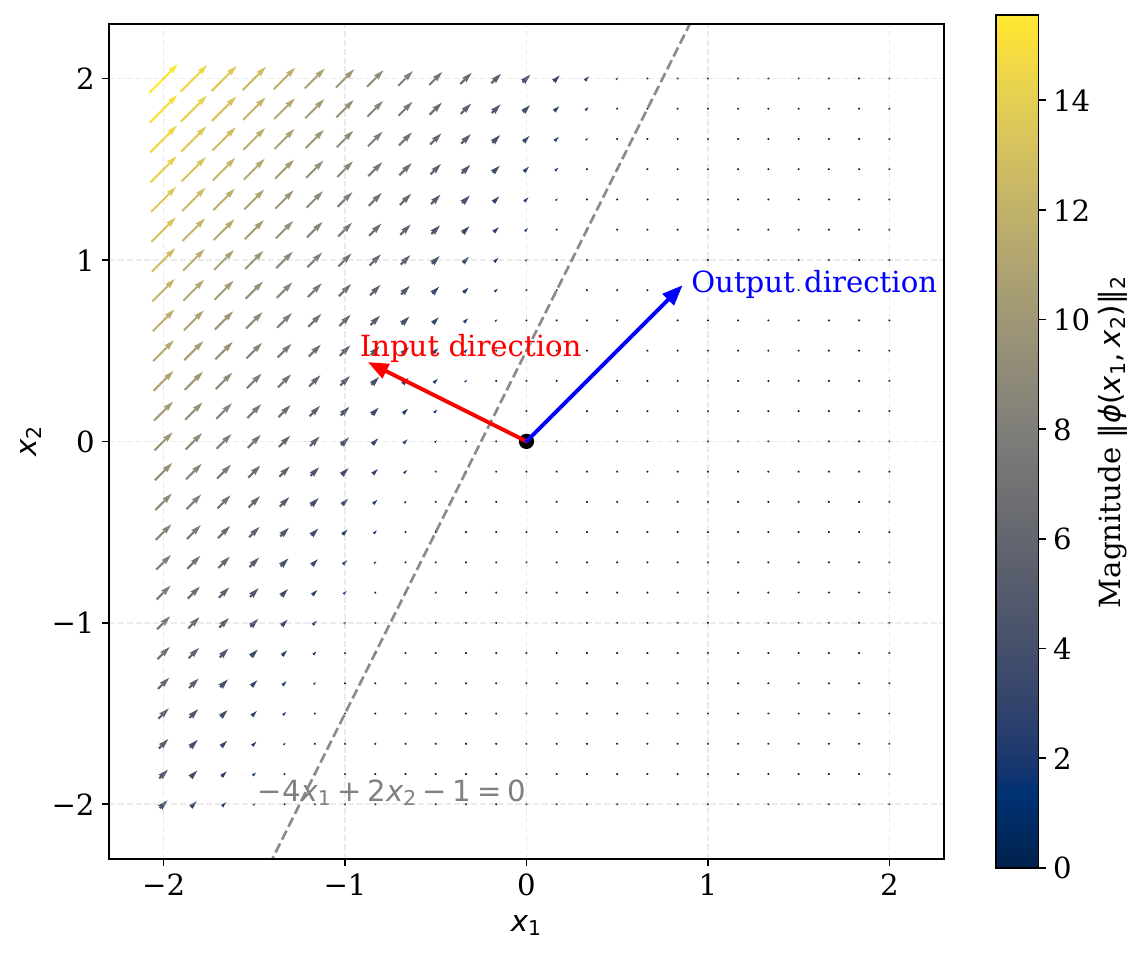}
        \caption{Visualization of the vector field defined by a single neural basis.}
        \label{fig:single_vec_field}
    \end{subfigure}
    \hfill
    \begin{subfigure}[t]{0.45\textwidth}
        \centering
        \includegraphics[width=\linewidth]
        {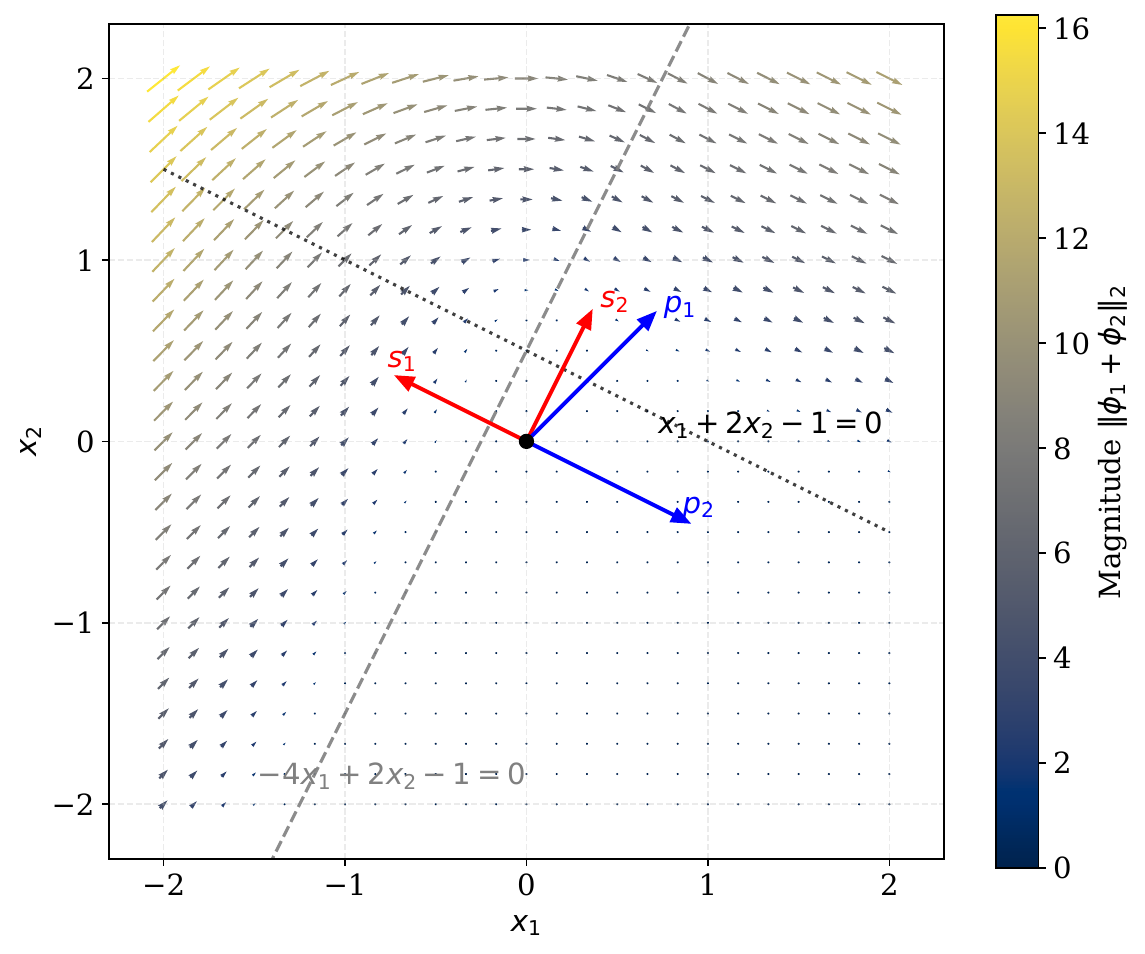}
        \caption{Visualization of a superposition of two vector fields induced by two neural bases.}
        \label{fig:two_vec_fields}
    \end{subfigure}
    \caption{Geometric interpretation of neural bases from the perspective of vector fields.
    \textbf{(a)} Visualization of a single neural basis. 
    The function is defined as $\phi(x)=\sigma(\langle x,s\rangle+b)p$,
    where $s=(-4,2)$, $b=-1$, $p=(1,1)$, and
    $\sigma$ is the GELU activation function~\citep{hendrycks2023gelu}.
    The red and blue arrows illustrate the directions of sensor $s$ and response $p$ in the input and output vector fields, respectively.
    The dashed line marks the activation boundary induced by the equation \(\langle x, s \rangle + b = 0\).
    The arrows of the vector field show the push-forward effect of the point on the grid.
    \textbf{(b)} A superposition of two vector fields, where the first neural basis is the same as that shown in \textbf{(a)} and the second basis has parameters \(s_2 = (1, 2)\), \(b_2 = -1\), and \(p_2 = (2,-1)\). Note that a single basis can only introduce one directional push-forward effort, whereas the combination of two bases can generate multiple directional effects, as exhibited by the rotational dynamics in \textbf{(b)}.
    }
    \label{fig:vec_field}
\end{figure}

\paragraph{Superposition of Vector Fields.}
We provide a geometric interpretation of the neural basis. 
Specifically, each neural basis can be viewed as a vector field from the input space to the output space. For the \(j\)-th neuron, as illustrated in Figure~\ref{fig:single_vec_field}, the sensor parameters \((s_j,b_j)\) determine the direction and bias of the input vector field, and the response vector \(p_j\) determines the direction of the push-forward in the output space. 
A bilinear structure is used to connect the nonlinear activation \(\sigma(\langle x,s_j\rangle+b_j)\) with the response vector \(p_j\). 
Consequently, a residual feedforward module can be interpreted as a \textit{superposition} of many such vector fields, as illustrated in Figure~\ref{fig:two_vec_fields}.

\paragraph{Projection Neural Bases.}
Beyond residual feedforward modules, the concept of neural bases can be extended to other types of neural network modules, such as linear projection modules. 
For such modules, we introduce a \textit{sensor--response duplex structure} that can parallelize the linear projection modules to form the corresponding \textit{projection neural bases}. 
Due to space limitations, the detailed formulation is provided in Appendix~\ref{sec:proj_nb}.

\subsection{Linear Reusable Neural Bases Architecture}
Figure~\ref{fig:lrnba_arch} illustrates the overall architecture.
We first introduce its two main components: reusable neural bases and a coefficient table.
We then describe the implementation details and present a sub-neuron reuse technique to further improve parameter efficiency.
Throughout this section, we consider a residual network comprising \(L\) stacked feedforward blocks.

\paragraph{Model Architecture.}
Essentially, our architecture is guided by two principles: \textit{neuron reuse} and \textit{linear combination}.
These principles give rise to two main components:
\begin{enumerate}
    \item \textbf{A collection of shared neural bases}, denoted by \(\{\phi_j\}_{j=1}^{m}\), which serve as shared basic nonlinear functions for constructing residual blocks.
    \item \textbf{A coefficient table}, denoted by \(C=(c_{j,l})\in\mathbb{R}^{m\times L}\), whose entries determine the contribution of each neural basis to each block.
\end{enumerate}
By using the above components, the residual block at depth \(l\) can be constructed as
\[
R_l(x) = \sum_{j=1}^{m} c_{j,l}\,\phi_j(x).
\]
This formulation reuses a shared collection of neural bases across depths while allowing different blocks to represent distinct transformations through their depth-specific coefficients.

Optionally, we can also add a shift table \(\Delta=(\delta_{j,l})\in\mathbb{R}^{m\times L}\) for the neural bases to provide depth-specific shifts and further increase their flexibility.
Here, we treat each coefficient \(c_{j,l}\) as an independent, unconstrained real-valued parameter.
Alternatively, constraints, such as non-negativity or binary values, may be imposed to control how neural bases contribute to each block, which may influence the topological connectivity of the network, such as sparse or dense structures.
We leave the investigation of these constraints to future work.

\paragraph{Remark.}
The approaches of \cite{savarese18parametersharing, wang2025basis} share a similar spirit with our framework by representing different layers as linear combinations of a shared set of bases with learned coefficients. However, there are two key distinctions. First, these approaches are primarily designed for CNN-based or attention-based architectures, whereas our framework focuses on FFN-based architectures. Second, they consider reusable transformations at the module or component level, while our framework introduces reuse at the finer-grained neuron level.


\begin{center}
    \begin{minipage}[t]{0.48\textwidth}
        \vspace{-0.3in}
        \begin{algorithm}[H]
            \caption{Forward pass across all depth blocks via reusable neural bases.}
            \label{alg:depth_forward}
            \begin{algorithmic}[1]
                \For{$l \gets 1$ \textbf{to} $L$}
                    \State $a \gets \sigma(Sx + b \textcolor{red}{+ \delta_l})$
                    \State \textcolor{red}{$a \gets a \odot (Mc_l)$}
                    \State $x \gets x + Pa$
                \EndFor
            \end{algorithmic}
        \end{algorithm}
    \end{minipage}
    \hfill
    \begin{minipage}[t]{0.48\textwidth}
        \vspace{-0.3in}
        \begin{algorithm}[H]
            \caption{Average accumulated gradients of reusable neural bases.}
            \label{alg:grad_avg}
            \begin{algorithmic}[1]
                \For{each shared parameter $p$}
                    \State $r_p \gets$ number of times $p$ is reused
                    \State Register a gradient hook on $p$:
                    \Statex \hspace{\algorithmicindent}
                        $\mathrm{grad} \mapsto \mathrm{grad}/r_p$
                \EndFor
            \end{algorithmic}
        \end{algorithm}
    \end{minipage}
\end{center}

\paragraph{Implementation Details.}
For initialization, the sensor parameters of the reusable neural bases are initialized using Kaiming initialization~\citep{he15init}, while the response parameters are zero-initialized to provide a neutral starting point for stable training, following the guidance of~\cite{Bachlechner21Rezero, zhang2019fixup}.
To accelerate the optimization of the coefficient table, we parameterize the effective coefficients as \(M c_l\) instead of \(c_l\) in the implementation, where \(M\) is a large constant (e.g., \(M=1000\) in our experiments). The parameters of the coefficient table are initialized from a Gaussian distribution \(\mathcal{N}(0, 1/\sqrt{M})\). 
For simplicity, the diagram still uses the effective coefficients \(c_l\).

Our training method mostly follows the standard gradient descent method. 
However, because the neural bases are reused across multiple blocks, their gradients are accumulated over all blocks for each optimization step.
To maintain the magnitude of gradient updates consistent with standard models, we rescale the accumulated gradients of the neural bases by the number of times they are reused.
The implementation details are summarized in Algorithm~\ref{alg:depth_forward} and~\ref{alg:grad_avg}.

\paragraph{Sub-Neuron Reuse.}
Beyond neuron-level reuse, we further consider reusing \textit{sub-neuron} components, such as sensors or responses, to improve parameter efficiency. 
One simple approach is to allow multiple neurons to share a single response vector, forming a more complex neuron unit with multiple sensors and a shared response, which we refer to as a \textit{multi-sensor neuron}. Networks employing activation functions such as Maxout~\citep{Goodfellow13Maxout} or SwiGLU~\citep{Shazeer20swiGLU} can be interpreted as examples constructed from such multi-sensor neurons. 
In addition to improving parameter efficiency, multi-sensor neurons often have larger activation regions, which may help mitigate the dead-neuron issue.

%% file: sec/disc.tex
\section{Discussion}
\label{sec:disc}
The neuron-reuse mechanism introduces several new benefits to the model, particularly in terms of network compression and parameter count. 
We discuss these benefits below, followed by a brief discussion of the limitations.


\paragraph{Network Compression.}
One important aspect of the proposed architecture is network compression. 
Compared with classical architectures with independent parameters, the neuron-reuse mechanism allows the same neural bases to be reused across depths. 
This reuse may enable the network to represent recurrent or repeated structural patterns more intrinsically and effectively, leading to more concise representations.
As suggested by the philosophy that compression leads to intelligence, representing a function with fewer parameters may also provide an implicit regularization effect. 
Therefore, from this perspective, the proposed architecture can offer improved generalization relative to classical architectures with fully independent parameters.

\paragraph{Parameter Count.}
In classical architectures with independent parameters, the total parameter count is mainly determined by three dimensions of the model: the embedding dimension, width (hidden dimension), and depth. Designing these models requires balancing these three dimensions under a memory budget. 
In contrast, our architecture reuses a shared collection of neurons across depths. 
As a result, the total parameter count is primarily determined by the embedding dimension and width, with less dependence on depth.\footnote{Although the coefficient table also introduces a parameter count that grows linearly with depth, this is negligible compared with that of fully independent modules.}
In practice, parameter count does not translate directly into memory usage, which also depends on implementation details and runtime requirements. 
Nevertheless, under the same parameter budget, our architecture enables the construction of substantially wider and deeper models.


\paragraph{Limitations.}
A primary limitation of our method is the increased computational cost compared with classical architectures, as each module must be constructed through a linear combination of neural bases. 
This additional computation may also lead to higher energy consumption. 

%% file: sec/exp.tex
\section{Experiments}
\label{sec:exp}
We evaluate the compression capability of the proposed architecture on two model types: FFN-based ResNets for regression and decoder-only Transformers for next-token prediction. 
Both models use GELU~\citep{hendrycks2023gelu} as the activation function.
For fair comparisons, all models use the same random seed and identical mini-batch sampling order throughout training. 
All experiments are conducted on a Linux workstation equipped with an Intel Core i7 CPU, 16 GB of RAM, and an NVIDIA RTX 3090 GPU.

\subsection{Evaluation of FFN-based ResNet models}
\label{sec:exp1}

\paragraph{Dataset and Model Architecture.}
The regression task is conducted on a synthetic dataset generated by the composite function
$f(x)=\cos\!\left(2\pi \cos(2\pi x)\right).$
We sample \(N=60{,}000\) points randomly on the interval \([-1,1]\) and split them into training and test sets using a \(3{:}1\) ratio.

The default architecture is an FFN-based residual network composed of a sequence of residual blocks, together with input and output linear projection modules. 
The three dimensions of the model are specified as follows: the embedding dimension is \(d=64\), the width is \(m=4d=256\) (i.e., the hidden dimension or number of neurons), and the depth is \(L=8\), where the depth counts only the FFN-based residual blocks.
To improve training stability, we use the parallelized linear projection modules introduced in Section~\ref{sec:proj_nb}, with their corresponding response parameters initialized to zero.

\paragraph{Training Setup.}
We use mean squared error (MSE) as the loss function and AdamW as the optimizer, with a learning rate of \(1\times10^{-5}\). All models are trained for \(4{,}096\) epochs with a batch size of \(512\). We report the training and test MSE losses, average training time, and visual comparisons between the model predictions and the ground-truth function.

\begin{figure}[ht]
\centering
\begin{subfigure}[t]{0.48\textwidth}
\centering
\includegraphics[width=\linewidth]
{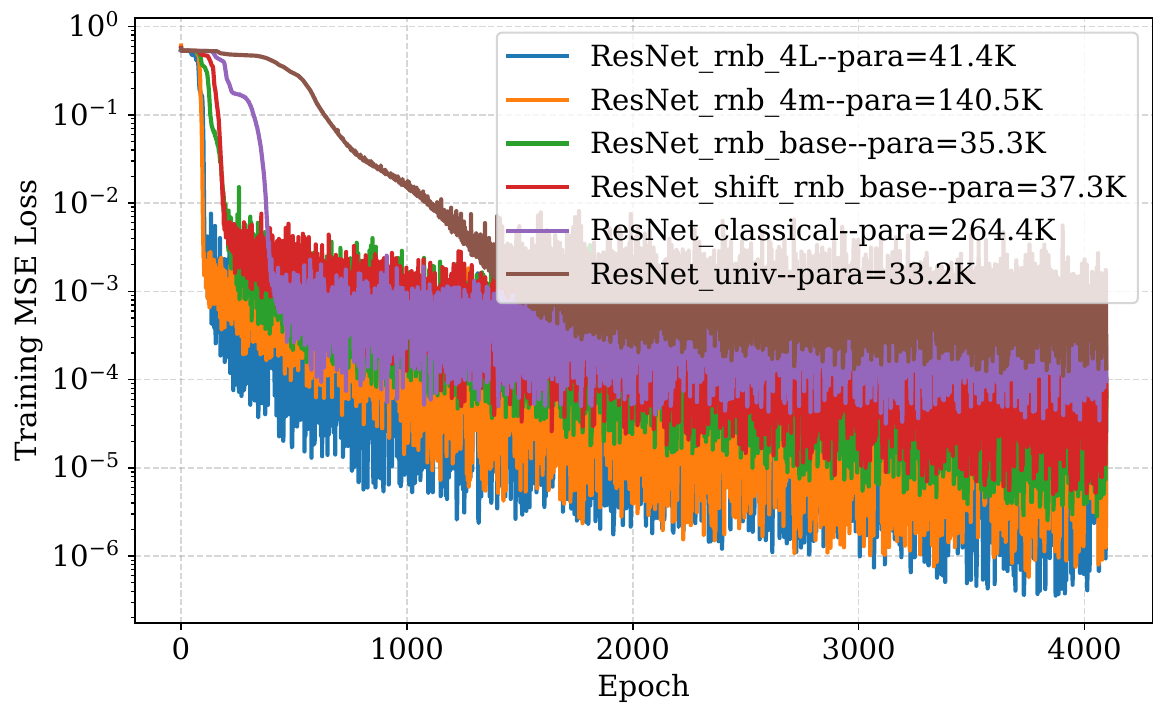}
\caption{Training MSE loss over epochs.}
\label{fig:regression_train_loss}
\end{subfigure}
\hfill
\begin{subfigure}[t]{0.48\textwidth}
\centering
\includegraphics[width=\linewidth]
{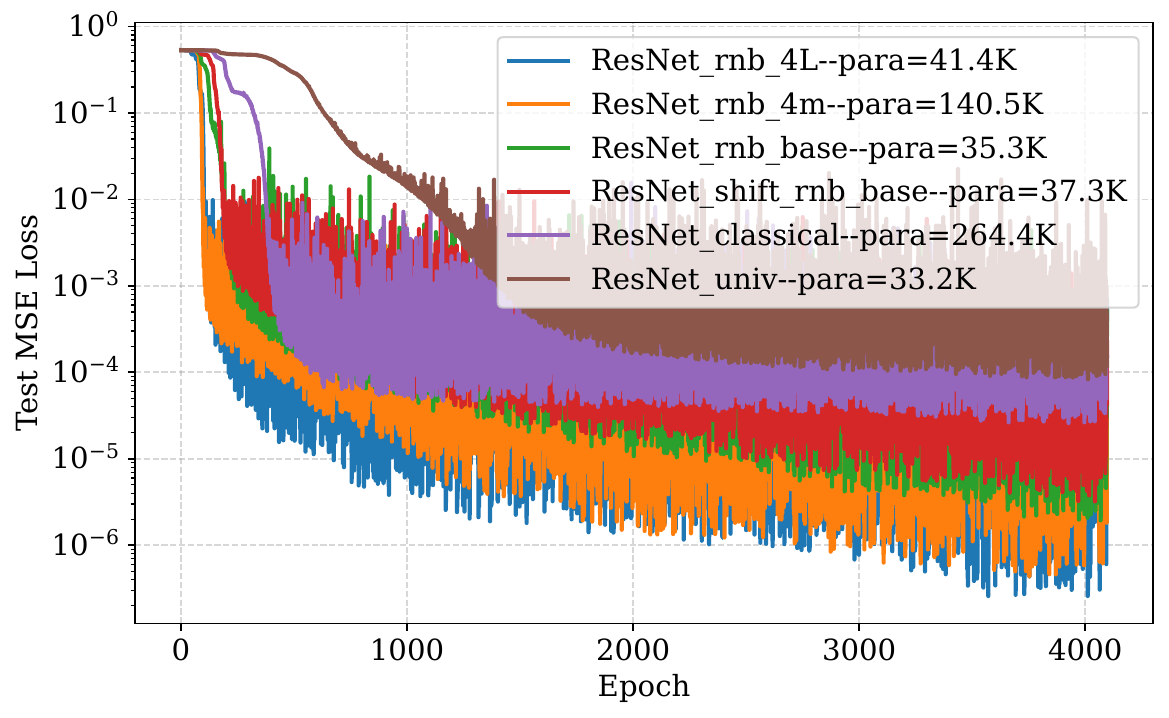}
\caption{Test MSE loss over epochs.}
\label{fig:regression_test_loss}
\end{subfigure}
\caption{
Evaluation of FFN-based ResNet models on a regression task.
\textbf{(a)} Training MSE loss over epochs. 
The proposed architectures achieve comparable or even lower final training loss than the classical ResNet and universal ResNet baselines, while using substantially fewer parameters.
\textbf{(b)} Test MSE loss over epochs. The test-set results exhibit trends consistent with those observed on the training set. In particular, the proposed models achieve lower test loss than the baselines, while using fewer parameters, suggesting that the neuron-reuse mechanism can yield more compact and parameter-efficient representations.
}
\label{fig:regression_task}
\end{figure}

\paragraph{Results.}
Figure~\ref{fig:regression_task}(a) shows that the proposed base architecture achieves a final loss comparable to that of the classical architecture while using only approximately \(1/7\) of its parameters. It also outperforms the universal architecture with a similar parameter count, in which the same residual module is repeatedly applied across all depths.

Furthermore, increasing the width by a factor of \(4\), denoted by \texttt{4m}, leads to a lower final loss while using only about \(1/2\) of the parameters of the classical architecture. 
Similarly, increasing the depth by a factor of \(4\), denoted by \texttt{4L}, also reduces the final loss while requiring only about \(1/6\) of the parameters of the classical architecture. 
These results demonstrate that both width and depth expansion can improve model performance under a reduced parameter budget. 
The depth expansion can further improve parameter efficiency, although it may incur a higher computational cost and longer training time.

Figure~\ref{fig:regression_task}(b) illustrates the generalization performance on the test set.
The test loss exhibits trends consistent with those observed on the training set. In particular, the proposed models achieve lower test loss while using fewer parameters than the classical ResNet baseline, suggesting that the neuron-reuse mechanism can  yield a more compact and parameter-efficient representation.

Due to space limits, results on training time and model predictions are provided in Appendix~\ref{sec:add_exp1}.

\subsection{Evaluation of Decoder-Only Transformer Models}

\paragraph{Dataset, Tokenization, and Model Architecture.}
We use the WikiText-2 dataset~\citep{merity2016pointer}, which contains approximately two million tokens and preserves the original case, punctuation, and document structure, for the next-token prediction task. The text is first tokenized using a byte-level BPE tokenizer with a vocabulary size of \(V=2^{14}=16{,}384\), including four special tokens: \texttt{UNK}, \texttt{PAD}, \texttt{EOS}, and \texttt{SOS}. We then concatenate all sentences into a single token sequence and randomly sample overlapping chunks of length \(T=128\).

For the model architecture, we use decoder-only Transformer models~\citep{Vaswani17attention, radford2019language}, including input and output parallelized linear projection blocks. 
The FFN-based modules share neural bases through the LRNB architecture.
The model configuration is specified as follows: embedding dimension \(d=512\), depth \(L=8\), and \(H=8\) attention heads, corresponding to a head dimension of \(64\). The hidden dimension of each FFN block is \(m=4d=2048\). No dropout is applied.

\paragraph{Model Variants and Baselines.}
The classical Transformer uses independently parameterized
blocks at each depth.
The fully shared baseline, denoted by \texttt{TF\_univ},
reuses a single Transformer block across all depths.
The FFN-shared baseline, denoted by \texttt{TF\_ffn\_univ},
reuses a single FFN module across all depths while retaining
independently parameterized attention modules.
The shifted LRNB variant, denoted by
\texttt{TF\_ffn\_shift\_rnb\_base}, augments the base LRNB
architecture with the learnable depth-specific shifts
$\delta_l$ introduced in Section~\ref{sec:method}.

\paragraph{Training Setup.}
We use cross-entropy (CE) as the loss function and AdamW as the optimizer, with a learning rate of \(1.5\times10^{-5}\). All models are trained for \(300\) epochs. Each epoch consists of \(4,096\) training batches with a batch size of \(16\), while evaluation is performed using \(128\) test batches. We report the evolution of the training and test CE losses throughout training, together with the average training time and total parameter count.

\begin{figure}[ht]
    \centering

    \begin{subfigure}[t]{0.48\textwidth}
        \centering
        \includegraphics[width=\linewidth]
        {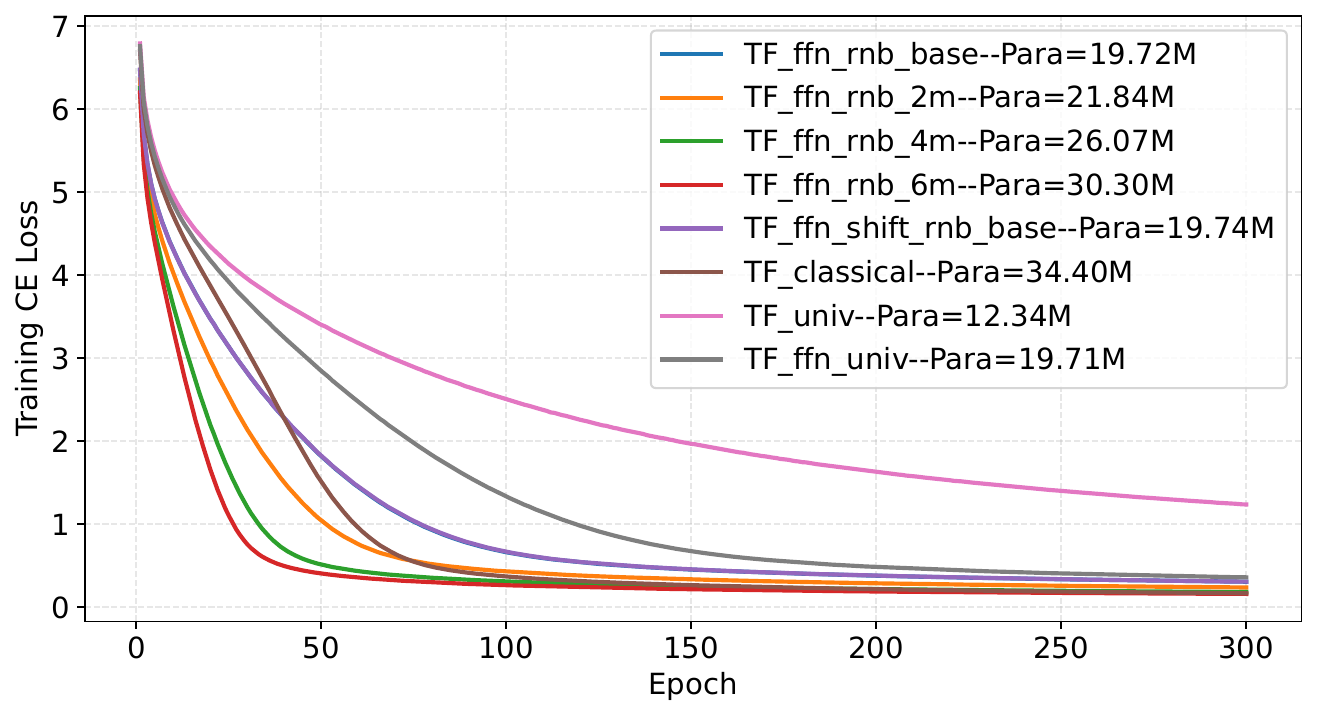}
        \caption{Training cross-entropy loss over epochs.}
        \label{fig:transformer_training_loss}
    \end{subfigure}
    \hfill
    \begin{subfigure}[t]{0.48\textwidth}
        \centering
        \includegraphics[width=\linewidth]
        {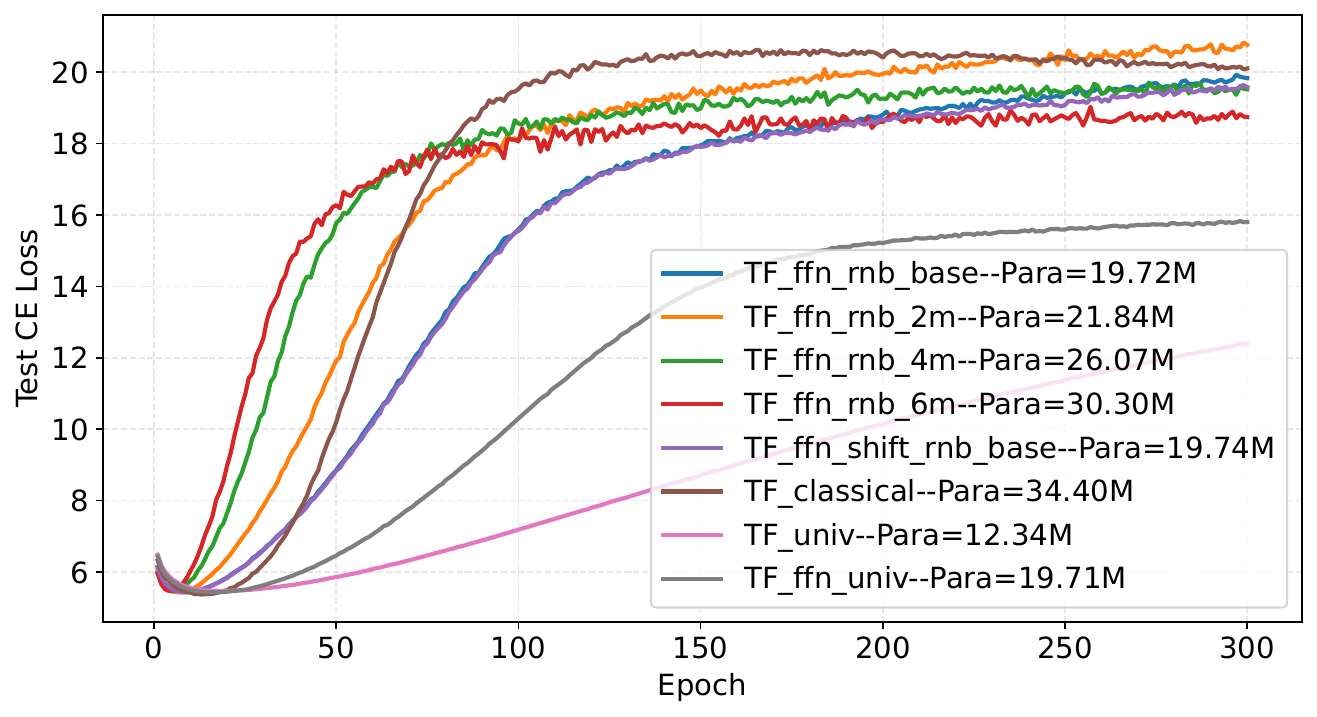}
        \caption{Test cross-entropy loss over epochs.}
        \label{fig:transformer_test_loss}
    \end{subfigure}
    \caption{
    Evaluation of decoder-only Transformer models on the next-token prediction task.
    \textbf{(a)} Training cross-entropy loss over epochs. Increasing the number of reusable neural bases from \texttt{2m} to \texttt{4m} and \texttt{6m} leads to lower final training loss. 
    In particular, the proposed architecture with an expansion factor of \(6\) (\texttt{6m}) achieves a lower final training loss than the classical Transformer while using approximately $12\%$ fewer parameters
($30.30$ million versus $34.40$ million).
    \textbf{(b)} Test cross-entropy loss over epochs. The classical Transformer yields a higher test CE loss than most LRNB models, with the exception of \texttt{2m}. 
    This observation is largely consistent with the philosophy that compression leads to intelligence---models with fewer parameters tend to generalize better on test data.
    }
    \label{fig:transformer_results}
\end{figure}

\vspace{-0.1in}
\paragraph{Results.}
Figure~\ref{fig:transformer_results}(a) shows the training cross-entropy loss over epochs. Increasing the number of reusable neural bases from \texttt{2m} to \texttt{4m} and \texttt{6m} leads to lower final training loss. In particular, the proposed architecture with an expansion factor of \(6\) (\texttt{6m}) outperforms the classical Transformer in final training loss while using approximately $12\%$ fewer parameters
($30.30$ million versus $34.40$ million).
These results demonstrate that the proposed reuse mechanism can improve parameter efficiency while maintaining stable training dynamics.

Figure~\ref{fig:transformer_results}(b) shows that the classical Transformer yields a higher test CE loss than most LRNB models, with the exception of \texttt{2m}.
This observation is consistent with the philosophy that compression is intelligence---the model with fewer parameters tends to achieve better generalization on test data. 

Average training times are provided in
Appendix~\ref{sec:add_exp2}.

%% file: sec/con.tex
\section{Conclusion}
\label{sec:conclusion}
In this paper, we introduced the Linear Reusable Neural Bases (LRNB) architecture, a new framework for reducing parameter redundancy and memory costs. 
By representing each feedforward module as a linear combination of reusable neural bases, the LRNB architecture integrates a parameter-reuse mechanism at the neuron level and provides the flexibility to construct substantially wider and deeper architectures under a fixed parameter budget.

Experiments on FFN-based ResNets and decoder-only Transformers further demonstrate that the proposed framework can outperform classical models with fully independent parameters in terms of both training and test loss, while using fewer parameters and maintaining stable training.

More broadly, integrating neuron reuse, either throughout the network or within selected modules, can yield more compact and parameter-efficient representations and may also improve generalization compared with classical models that use fully independent parameters.
These findings highlight neural-basis reuse as a promising direction for memory-efficient deep learning and motivate further investigation into more general reuse structures, coefficient constraints, pruning strategies, and large-scale architectures.

\newpage
\section*{Ethics Statement}
This work examines parameter sharing in neural networks and evaluates the approach on a synthetic dataset and a language modeling task based on the WikiText-2 dataset.
It does not involve human-subject research or the collection of personal data. 
Models trained on existing text corpora may inherit biases or produce inaccurate content; the proposed architecture does not directly address these limitations.

\section*{Reproducibility Statement}
The paper describes the neural-basis parameterization, forward computation, initialization, and gradient-rescaling procedure.
The experimental sections specify the datasets, model dimensions, optimizer, learning rates, batch sizes, training duration, and hardware.
Additional experimental results are provided in the appendices. 
Comparisons use a common random seed and identical mini-batch sampling order to control part of the experimental variation. 
These controls do not establish robustness across random seeds, which remains a limitation of the reported evaluation.

\section*{AI Use Statement}
Generative AI tools, including ChatGPT, Claude, and DeepSeek, were used to assist with manuscript proofreading, wording revisions, LaTeX troubleshooting, and checking mathematical notation and consistency between experimental descriptions and figures. 
These tools were also used to improve clarity and readability, assist with manuscript organization, summarize and identify relevant literature, and aid in reference formatting and related writing tasks. 
We did not use generative AI to directly generate the experimental datasets used in this work. 
Translation and qualitative or thematic data analysis were not applicable to this study.

All AI-assisted content was reviewed by the authors. 
In particular, mathematical definitions and derivations were independently checked by the authors; AI-assisted code was reviewed, executed, and tested; numerical results were obtained from the authors' experimental implementations; and references and literature-related information were checked against the original sources where appropriate. Generative AI was not treated as an authoritative source for mathematical or scientific claims.

%% file: sec/add_nbs.tex
\newpage

\section{Parallelized Projection Modules and Projection Neural Bases}
\label{sec:proj_nb}

In this section, we provide a more detailed formulation of \textit{parallelized projection modules} and \textit{projection neural bases}, together with their decomposition and geometric interpretation.

\paragraph{Sensor--Response Duplex Structure.}
To parallelize linear projection modules, we represent each projection as the sum of a fixed linear projection and a \textit{sensor--response duplex structure}, with the identity function serving as the activation function.
Since both the fixed branch and the duplex branch are affine, their sum is still an affine transformation.

Specifically, we adopt the same sensor--response notation used for the FFN-based ResNet. Let \((S,b)\) denote the sensor parameters and \(P\) the response parameters. The linear projection is written as
$$
\text{Proj}(x) + P(Sx+b),
$$
where $\text{Proj}(x)$ represents the linear projection layer from input space to output space with frozen parameters, using the Kaiming initialization~\citep{he15init}.

Let \(m\) denote the hidden dimension of the block, and write the rows of \(S\), the entries of \(b\), and the columns of \(P\) as
$$
S^\top=[s_1,s_2,\ldots,s_m],
\qquad
b^\top=[b_1,b_2,\ldots,b_m],
\qquad
P=[p_1,p_2,\ldots,p_m],
$$

where \(s_j^\top\) corresponds to the \(j\)-th row of \(S\), \(b_j\) is the \(j\)-th entry of \(b\), and \(p_j\) is the \(j\)-th column of \(P\).

For each \(j=1,2,\ldots,m\), we define
$$
\phi_j(x)
:=
(\langle x,s_j\rangle+b_j)p_j,
$$
where the identity function replaces the nonlinear activation function \(\sigma\) for the linear setting.

\paragraph{Decomposition.}
The duplex structure can then be decomposed as a sum of these elementary functions:
\begin{equation*}
P(Sx+b)
=
\sum_{j=1}^{m}
(\langle x,s_j\rangle+b_j)p_j
=
\sum_{j=1}^{m}\phi_j(x).
\end{equation*}
This formulation shows that the duplex structure of the projection block can be expressed as a superposition of elementary functions \(\{\phi_j\}_{j=1}^{m}\).
Based on this decomposition, we refer to each \(\phi_j\) as a \textit{projection neuron}, or equivalently a \textit{projection neural basis}. We similarly refer to \((s_j,b_j)\) as the \textit{sensor} of \(\phi_j\) and \(p_j\) as its \textit{response}.

A useful interpretation is that each projection neural basis corresponds to an affine transformation whose linear component has rank at most one.
Indeed,

$$
\phi_j(x)
=
(\langle x,s_j\rangle+b_j)p_j
=
p_j s_j^\top x+p_j b_j,
$$

where \(p_j s_j^\top\) is a matrix of rank at most one and \(p_j b_j\) is the corresponding bias vector. 
Thus, the trainable duplex component can be viewed as a superposition of multiple low-rank affine transformations.

\begin{figure}[ht]
\centering
\begin{subfigure}[t]{0.45\textwidth}
    \centering
    \includegraphics[width=\linewidth]
    {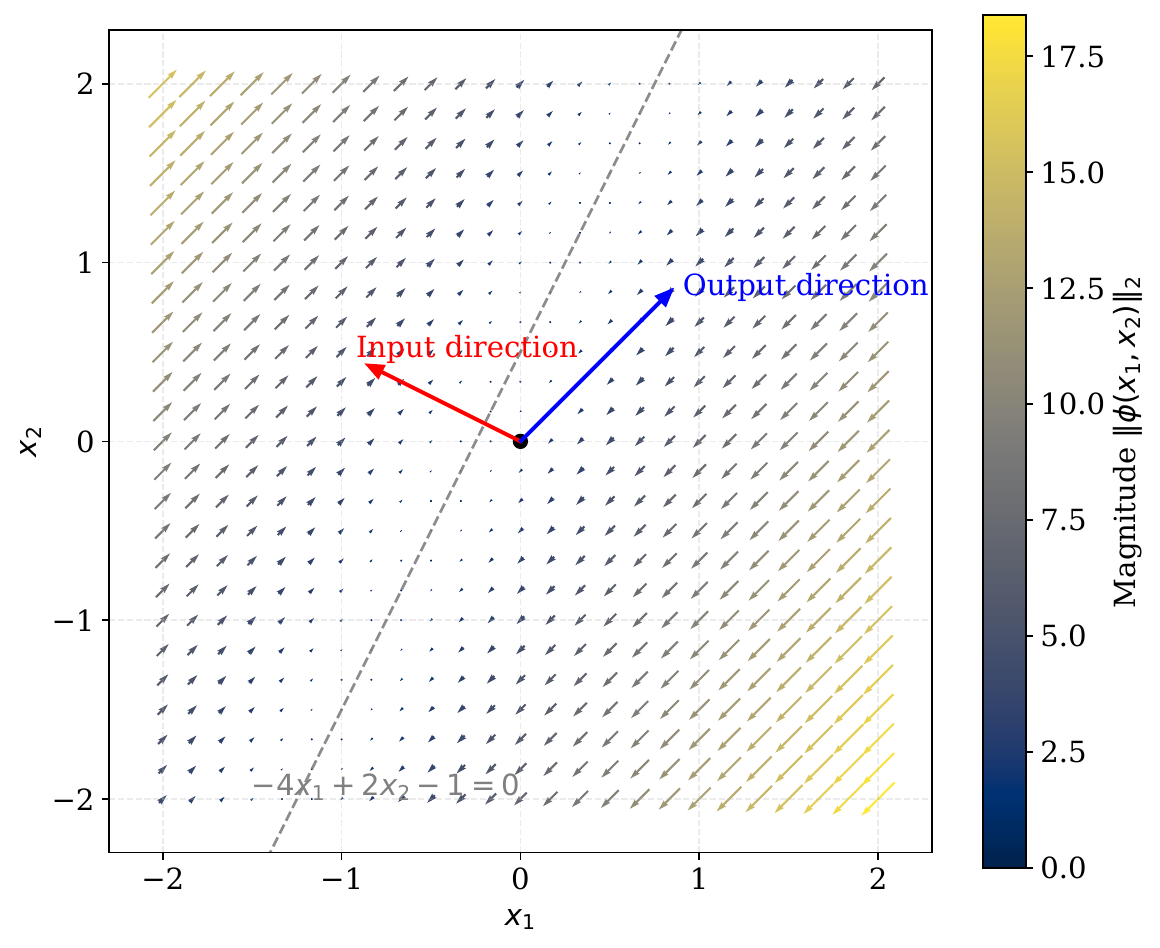}
    \caption{Vector field induced by a single projection neural basis.}
    \label{fig:single_vec_field_proj}
\end{subfigure}
\hfill
\begin{subfigure}[t]{0.45\textwidth}
    \centering
    \includegraphics[width=\linewidth]
    {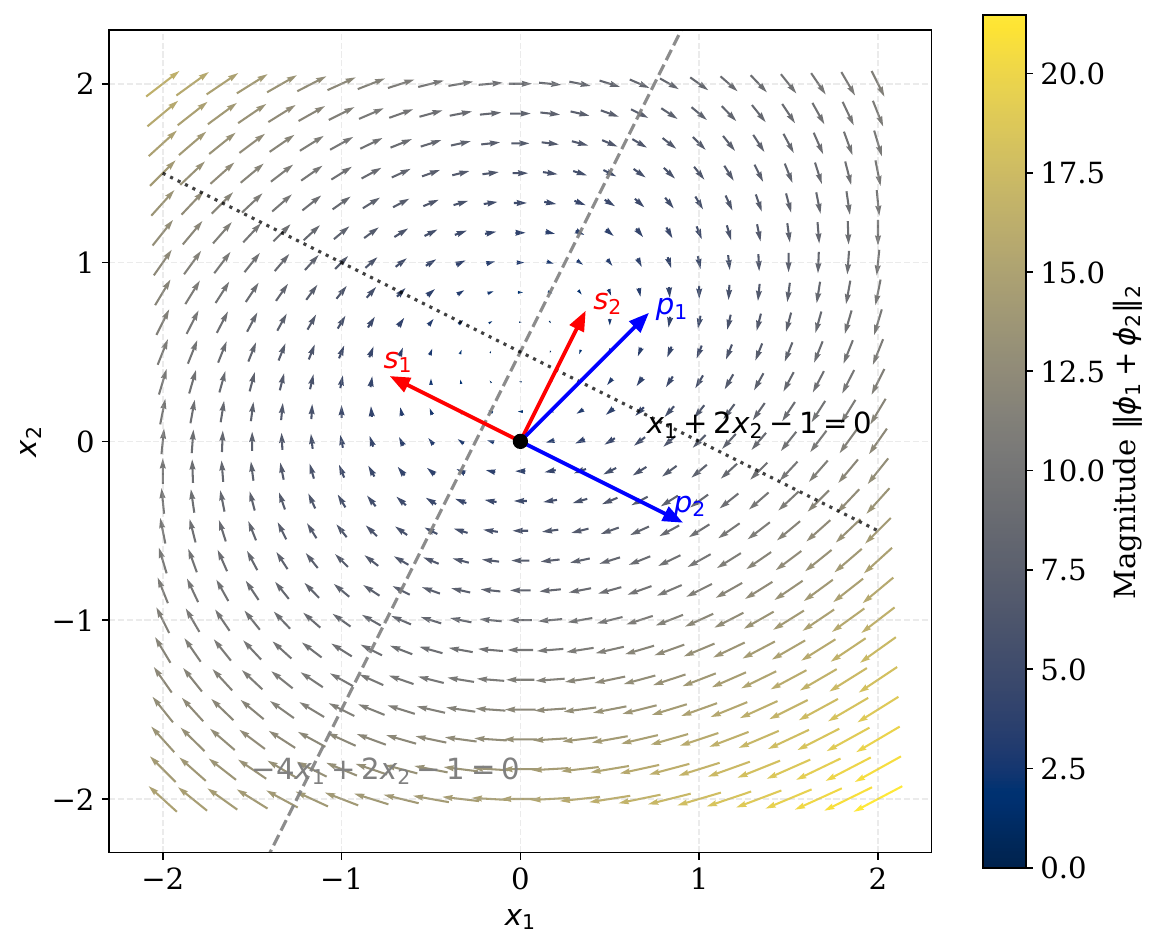}
    \caption{Vector field induced by the superposition of two projection neural bases.}
    \label{fig:two_vec_fields_proj}
\end{subfigure}
\caption{
Geometric interpretation of projection neural bases as vector fields.
\textbf{(a)} A single projection neural basis
\(\phi(x)=(\langle x,s\rangle+b)p\), with
\(s=(-4,2)\), \(b=-1\), and \(p=(1,1)\).
The red and blue arrows illustrate the sensor and response directions, respectively, while the dashed line indicates the zero line \(\langle x,s\rangle+b=0\).
The grid arrows show the output vectors induced at different input locations.
\textbf{(b)} Superposition of two projection neural bases. The first basis is identical to that in \textbf{(a)}, while the second is defined by
\(s_2=(1,2)\), \(b_2=-1\), and \(p_2=(2,-1)\).
While a single basis generates vectors along a fixed response direction, the superposition of multiple bases can produce more vector-field structures, such as rotational patterns.
}
\label{fig:vec_field_proj}
\end{figure}

\paragraph{Geometric Interpretation.}
We further provide a geometric interpretation of projection neural bases from the perspective of vector fields. 
We use the same parameter values as in the earlier vector-field visualizations shown in Figure~\ref{fig:vec_field}.

For the \(j\)-th projection neuron, as illustrated in Figure~\ref{fig:single_vec_field_proj}, the sensor vector \(s_j\) and scalar bias \(b_j\) determine how the input position \(x\) is mapped to the scalar response
$$
\langle x,s_j\rangle+b_j,
$$
while the vector \(p_j\) determines the direction of the resulting output vector. Their interaction therefore produces a vector field of the form
$$
\phi_j(x)
=
(\langle x,s_j\rangle+b_j)p_j.
$$
Consequently, the trainable duplex component can be interpreted as a superposition of such vector fields, as illustrated in Figure~\ref{fig:two_vec_fields_proj}.

%% file: sec/add_exps.tex
\newpage
\section{Additional Experimental Results of FFN-Based ResNets}
\label{sec:add_exp1}

\begin{figure}[ht]
\centering
\begin{subfigure}[t]{0.5\textwidth}
\centering
\includegraphics[width=\linewidth]{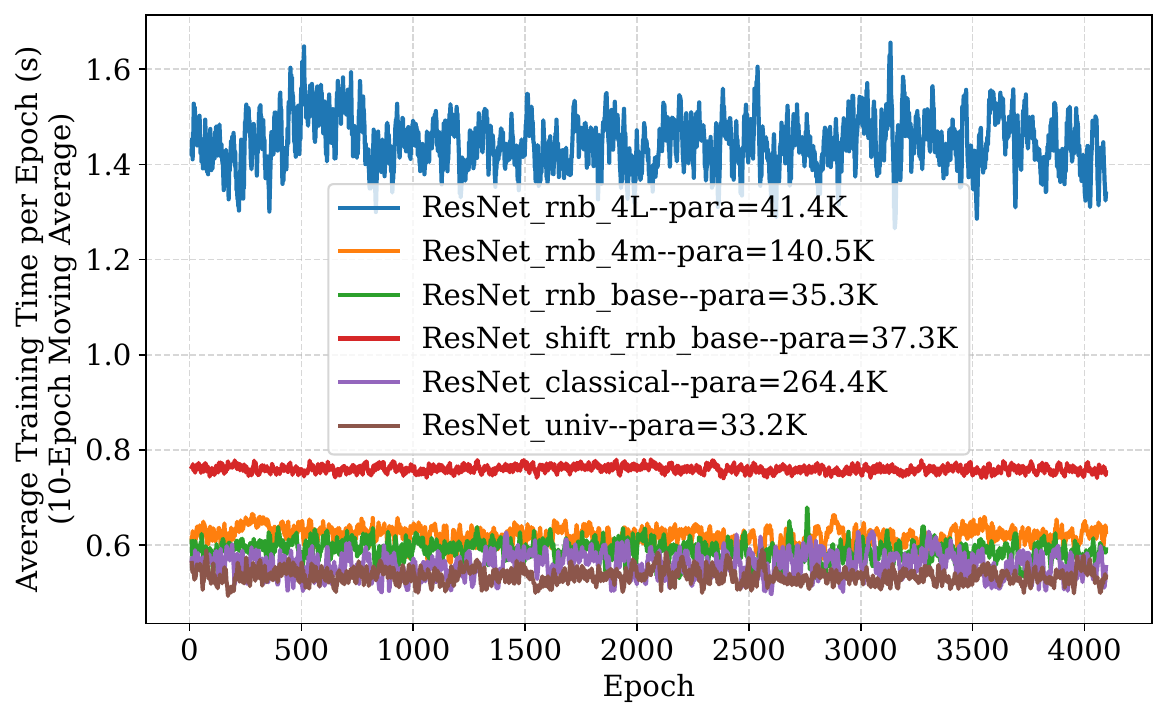}
\caption{Smoothed training time over epochs.}
\label{fig:test_mse_app}
\end{subfigure}
\hfill
\begin{subfigure}[t]{0.465\textwidth}
\centering
\includegraphics[width=\linewidth]{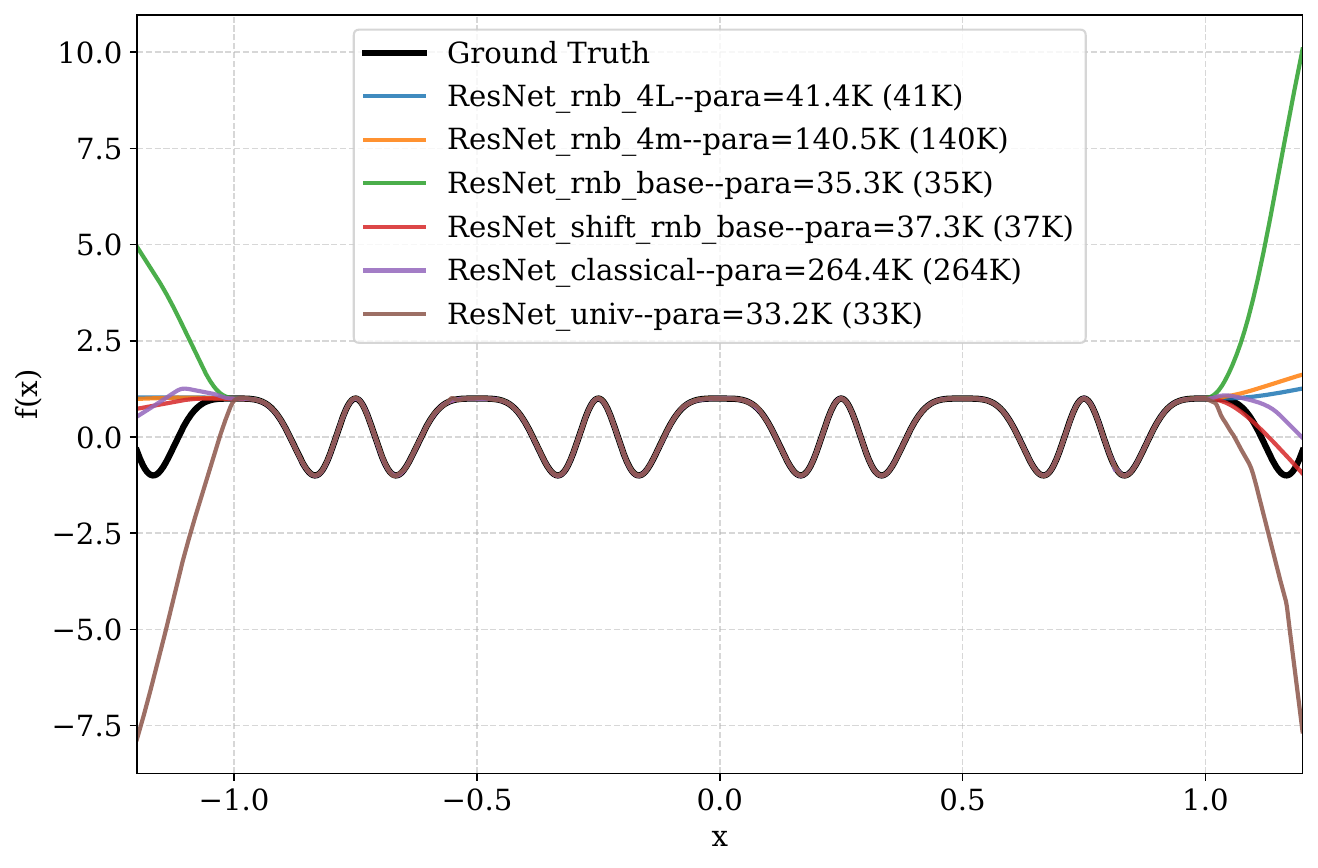}
\caption{Model predictions compared with the ground-truth function over
$[-1.2,1.2]$.}
\label{fig:pred_vs_gt_app}
\end{subfigure}
\caption{
Additional evaluation of the FFN-based ResNet models on the nonlinear regression task.
\textbf{(a)} Smoothed training time over epochs. The proposed architecture introduces a modest computational overhead due to the dynamic construction of residual blocks from reusable neural bases and depth-dependent coefficients.
\textbf{(b)} Model predictions compared with the ground-truth function over
$[-1.2,1.2]$. Training and test samples are drawn from $[-1,1]$;
predictions outside this interval illustrate extrapolation.
All evaluated models closely approximate the target function over $[-1,1]$.
}
\label{fig:test_and_viz}
\end{figure}

\paragraph{Results.}
Figure~\ref{fig:test_and_viz}(a) shows the smoothed training time over epochs. The proposed architecture introduces an additional training-time overhead because its residual blocks are dynamically constructed from reusable neural bases and depth-dependent coefficients. Figure~\ref{fig:test_and_viz}(b) compares the model predictions with the ground-truth function. 
All models provide accurate approximations to the target function within $[-1, 1]$.

\newpage
\section{Additional Experimental Results of Transformer models}
\label{sec:add_exp2}

\begin{figure}[ht]
    \centering

    \includegraphics[width=0.90\linewidth]{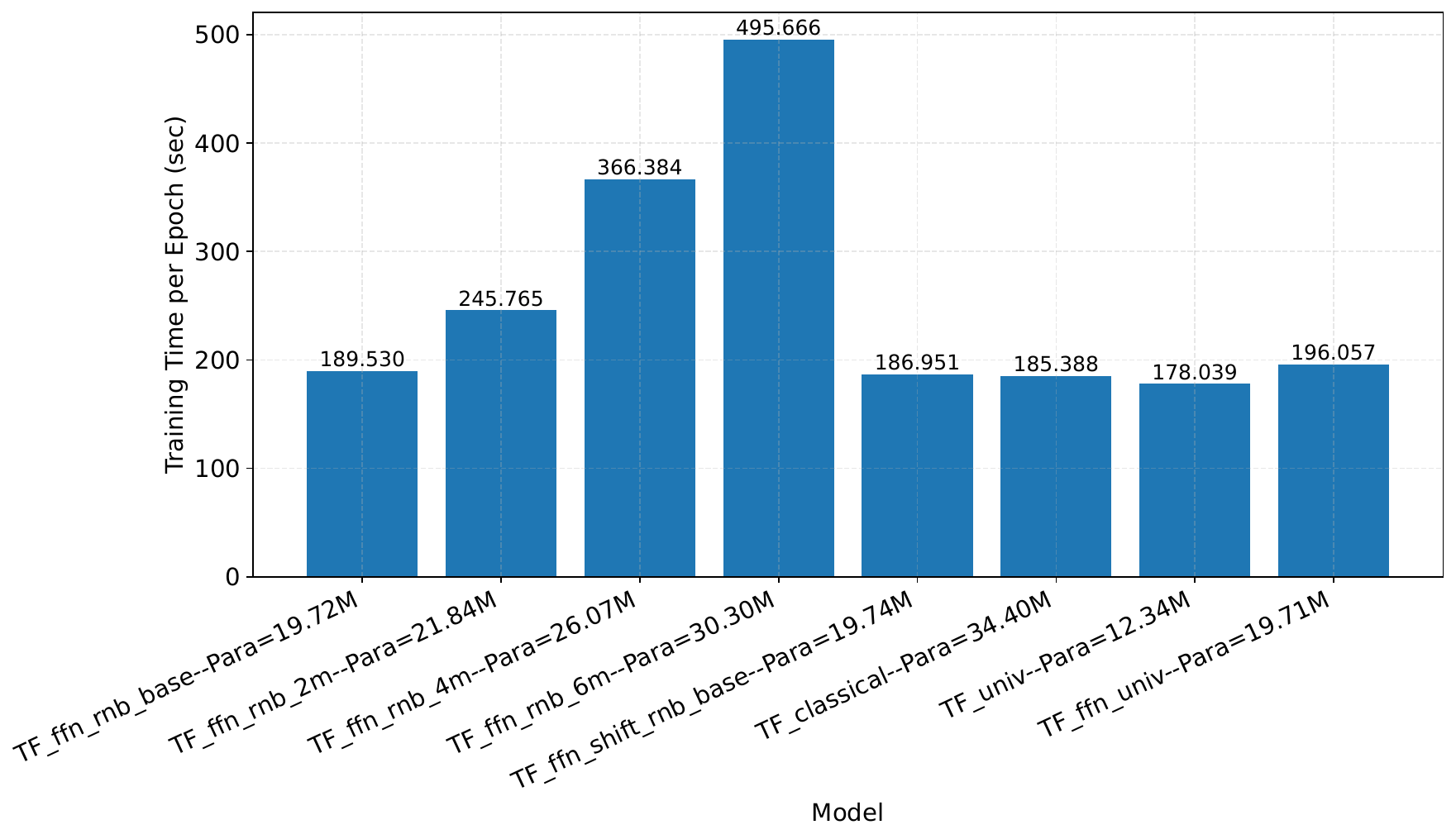}

    \caption{
    Average training time per epoch.
    The LRNB base model has a training time per epoch close to that of the classical baseline (189.5 s versus 185.4 s). Increasing the reusable FFN width raises training time, reaching 495.7 s for the 6m variant. These measurements illustrate the parameter–computation trade-off; inference latency and the hardware causes of the overhead were not evaluated here.
    }
    \label{fig:TF_time_eval}
\end{figure}

\paragraph{Results.}
Figure~\ref{fig:TF_time_eval} shows the average training time per epoch.
The LRNB base model has a training time per epoch close to that of the classical baseline (189.5 s versus 185.4 s). Increasing the reusable FFN width raises training time, reaching 495.7 s for the 6m variant. These measurements illustrate the parameter–computation trade-off; inference latency and the hardware causes of the overhead were not evaluated here.